\documentclass{article}
\usepackage{algorithm}
\usepackage{algorithmic}

\usepackage{amsmath}
\usepackage[preprint]{corl_2024} 

\newcommand{\name}{VLMnav}
\newcommand{\baseline}{Ours w/o nav}

\usepackage{wrapfig}
\usepackage{tabularx}
\usepackage[table]{xcolor}
\usepackage{changepage}
\usepackage{amsmath}
\usepackage{amssymb}
\usepackage{graphicx}
\usepackage{titlesec}
\titlespacing*{\subsubsection}{0pt}{*0.5}{*0.2}

\title{End-to-End Navigation with Vision-Language Models: Transforming Spatial Reasoning into Question-Answering}

\author{
  \textbf{Dylan Goetting}\\
  University of California Berkeley
  \\
  \texttt{dylangoetting@berkeley.edu} \\
  \AND
  \textbf{Himanshu Gaurav Singh} \\
    University of California Berkeley
 \\
  \texttt{himanshu\_singh@berkeley.edu} \\
  \and
  \textbf{Antonio Loquercio} \\
  University of Pennsylvania \\
  \texttt{aloque@seas.upenn.edu} \\
}

\begin{document}
\maketitle


\begin{abstract}
We present \name, an embodied framework to transform a Vision-Language Model (VLM) into an end-to-end navigation policy. In contrast to prior work, we do not rely on a separation between perception, planning, and control; instead, we use a VLM to directly select actions in one step. Surprisingly, we find that a VLM can be used as an end-to-end policy zero-shot, i.e., without any fine-tuning or exposure to navigation data. This makes our approach open-ended and generalizable to any downstream navigation task. 
    We run an extensive study to evaluate the performance of our approach in comparison to baseline prompting methods. In addition, we perform a design analysis to understand the most impactful design decisions. Visual examples and code for our project can be found at \href{https://jirl-upenn.github.io/VLMnav/}{jirl-upenn.github.io/VLMnav/}.

\end{abstract}


\vspace{30pt}

\section{Introduction}

The ability to navigate effectively within an environment to achieve a goal is a hallmark of physical intelligence. Spatial memory, along with more advanced forms of spatial cognition, is believed to have begun evolving early in the history of land animals and advanced vertebrates, likely between 400 and 200 million years ago~\cite{Muzio_Bingman_2022}. Because this ability has evolved over such a long period, it feels almost instinctual and trivial to humans. However, navigation is, in reality, a highly complex problem. It requires the coordination of low-level planning to avoid obstacles alongside high-level reasoning to interpret the environment’s semantics and explore the directions that are most likely to get the agent to achieve their goals.

A significant portion of the navigation problem appears to involve cognitive processes similar to those required for answering long-context image and video questions, an area where contemporary vision-language models (VLMs) excel~\cite{openai2024gpt4technicalreport,geminiteam2024gemini15unlockingmultimodal}. However, when naively applied to navigation tasks, these models face clear limitations. Specifically, when given a task description concatenated with an observation-action history, VLMs often struggle to produce fine-grained spatial outputs to avoid obstacles and fail to effectively utilize their long-context reasoning capabilities to support effective navigation~\cite{ramakrishnan2024doesspatialcognitionemerge, google2024pivot, rahmanzadehgervi2024visionlanguagemodelsblind}.

 \begin{figure}[h]
    \centering 
    \includegraphics[width=1\textwidth]{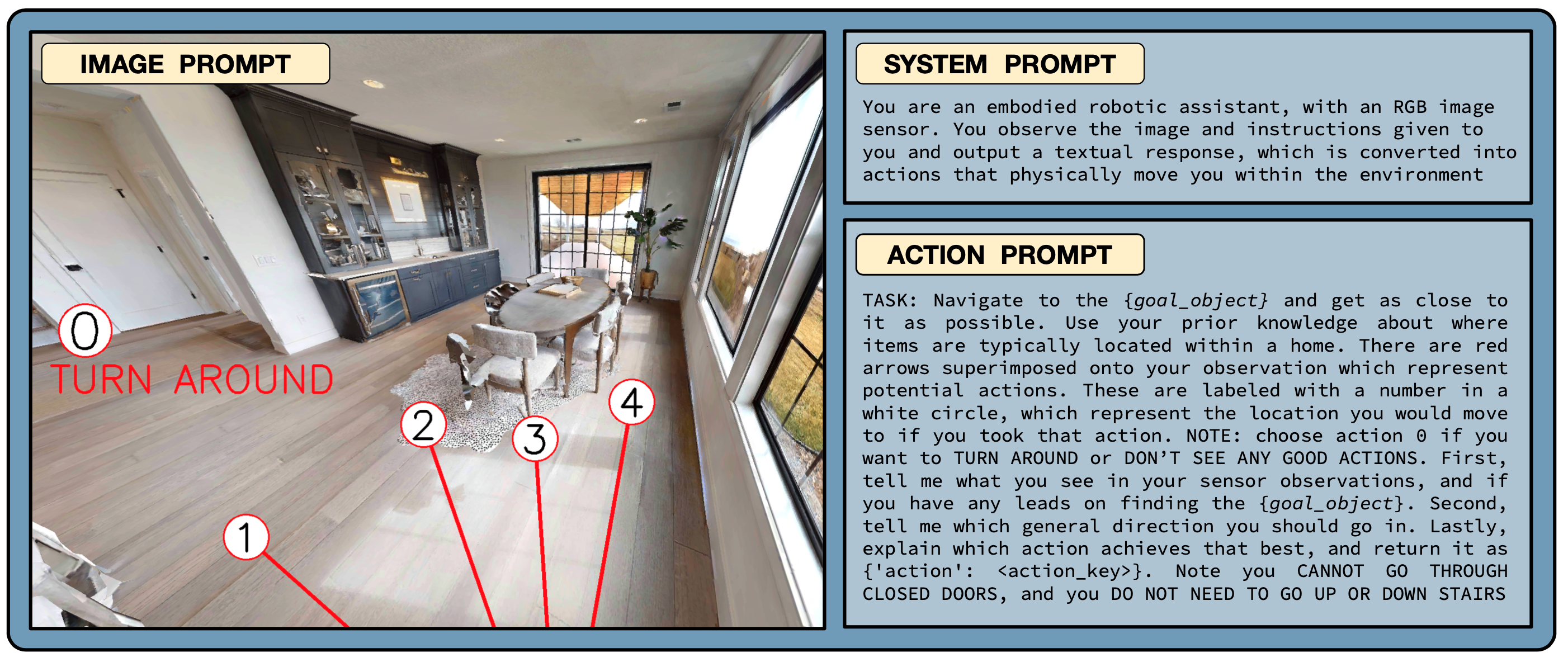}  
    \caption{\footnotesize The full action prompt for \name\ consists of three parts: A system prompt to describe the embodiment, an action prompt to describe the task, the potential actions, and the output instruction, and an image prompt showing the current observation along with the annotated actions}
    \label{fig:1}
\end{figure}
To address these challenges, previous work has included VLMs as a component of a modular system to perform high-level reasoning and recognition tasks. The systems generally contain an explicit 3D mapping module and a planner to deal with the more embodied part of the task, e.g., motion and exploration~\cite{kim2024openvla, majumdar2022zson, gadre2023cows, yu2023l3mvn, kuang2024openfmnavopensetzeroshotobject}.
While modularity has the advantage of utilizing each component only for the sub-task it excels at, it comes at the disadvantage of system complexity and task specialization. 

In this work, we show that an off-the-shelf VLM can be used as a zero-shot and end-to-end language-conditioned navigation policy. The key idea to achieve this goal is transforming the navigation problem into something VLMs excel at: \emph{answering a question about an image}. 

To do so, we develop a novel prompting strategy that enables VLMs to explicitly consider the problem of exploration and obstacle avoidance. This prompting is general, in the sense that it can be used for any vision-based navigation task.

Compared to prior approaches, we do not employ modality-specific experts ~\cite{exploreeqa2024, yu2023l3mvn, shah2023lfg}, do not train any domain-specific models \cite{zhang2024navid, spoc2023} and do not assume access to probabilities from the models \cite{exploreeqa2024, yu2023l3mvn}. 

We evaluate our approach on established benchmarks for embodied navigation \cite{habitatchallenge2022, khanna2024goatbench}, where results confirm that our method significantly improves navigation performance compared to existing prompting methods. Finally, we draw design insights from ablation experiments over several components of our embodied VLM framework. 

\section{Related Work}
\label{sec:citations}

The most common approach for learning an end-to-end navigation policy involves training a model from scratch using offline datasets~\cite{ving,shah2023rapidexplorationopenworldnavigation,chang2023goat,shah2023vintfoundationmodelvisual,shah2023gnmgeneralnavigationmodel}. However, collecting large-scale navigation data is challenging, and as a result, these models often struggle to generalize to novel tasks or out-of-distribution environments.

An alternative approach to enhance generalization is fine-tuning existing vision-language models (VLMs) with robot-specific data~\cite{brohan2022rt, brohan2023rt, kim2024openvla, zhang2024navid}. Although this method can lead to more robust end-to-end policies, fine-tuning may destroy features not present in the fine-tuning dataset, ultimately limiting the model's generalization ability.

An alternate line of work focuses on using these models zero-shot \cite{kuang2024openfmnavopensetzeroshotobject, zhou2023navgptexplicitreasoningvisionandlanguage,yu2023l3mvn,shah2023lfg, exploreeqa2024, gadre2023cows, google2024pivot}, by prompting them such that the responses align with task specifications. For instance, \cite{gadre2023cows, chang2023goat} use CLIP or DETIC features to align visual observations to language goals, build a semantic map of the environment, and use traditional methods for planning.  
Other works design specific modules to handle the task of exploration \cite{shah2023lfg, exploreeqa2024, kuang2024openfmnavopensetzeroshotobject,topiwala2018frontierbasedexplorationautonomous}. These systems often require an estimation of confidence to know when to stop exploring, which is commonly done using token or object probabilities \cite{exploreeqa2024, yu2023l3mvn}.
In addition, many of these approaches also use low-level navigation modules, which abstract away the action choices to a pre-trained point-to-point policy such as the Fast Marching Method \cite{chang2023goat, gadre2023cows, shah2023lfg, kuang2024openfmnavopensetzeroshotobject, yu2023l3mvn}.

\begin{figure}[t]
    \centering 
    \includegraphics[width=1\textwidth]{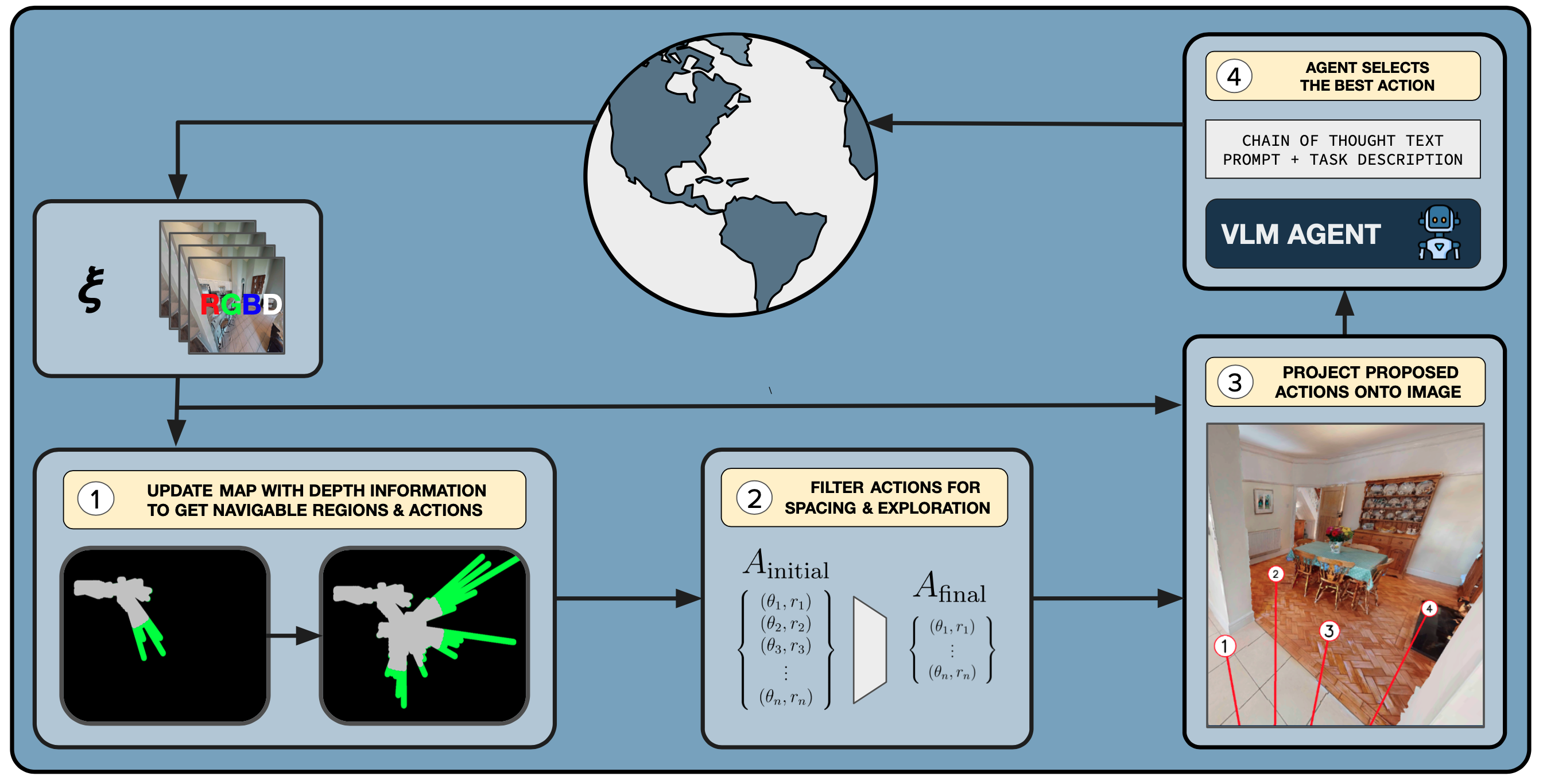}  
    \caption{\footnotesize\textbf{Approach:} Our method is made up of four key components: (i) \textit{Navigability}, which determines locations the agent can actually move to, and updates the voxel map accordingly. An example update step to the map shows the marking of new area as explored (gray) or unexplored (green). (ii)\textit{ Action Proposer}, which refines a set of final actions according to spacing and exploration. (iii) \textit{Projection}, which visually annotates the image with actions. (iv) \textit{Prompting}, which constructs a detailed chain-of-thought prompt to select an action. 
    \label{fig:overview}}
\end{figure}
\textbf{Visual Prompting Methods}: To enhance the task-specific performance of VLMs, recent work has involved physically modifying images before passing them to the VLM. Examples include \cite{shtedritski2023doesclipknowred}, which annotates images to help recognize spatial concepts.
\cite{yang2023setofmark} introduces \textit{set-of-mark}, which assigns unique labels to objects in an image and references these labels in the textual prompt to the VLM. This visual enhancement significantly improves performance on tasks requiring visual grounding. Building on this, \cite{koh2024visualwebarenaevaluatingmultimodalagents, yan2023gpt4vwonderlandlargemultimodal} apply similar visual prompting methods to the task of web navigation and show VLMs are able to complete such tasks zero shot. 

\textbf{Prompting VLMs for Embodied Navigation}: CoNVOI \cite{sathyamoorthy2024convoicontextawarenavigationusing} overlays numerical markers on an image and prompts the VLM to output a sequence of these markers in alignment with contextual cues (e.g., \textit{stay on the pavement}), which is used as a navigation path. Unlike our work, they (i) rely on a low-level planner for obstacle avoidance rather than using the VLM’s outputs directly as navigational actions, and (ii) do not leverage the VLM to guide the agent toward a specific goal location. 
PIVOT~\cite{google2024pivot}, introduces a visual prompting method that is most similar to ours. They approach the navigation problem by representing one-step actions as arrows pointing to labeled circles on an image. At each step, actions are sampled from an isotropic Gaussian distribution, with the mean and variance iteratively updated based on feedback from the VLM. The final action is selected after refining the distribution. While PIVOT is capable of handling various real-world navigation and manipulation tasks, it has two significant drawbacks: (i) it does not incorporate depth information to assess the feasibility of action proposals, leading to less efficient movement; and (ii) it requires many VLM calls to select a single action, resulting in higher computational costs and latency.

\section{Overview}
\label{sec:overview}

We present \name, designed as a navigation system that takes as input goal $\mathcal{G}$, which can be specified in language or an image, RGB-D image $I$, pose $\xi$, and subsequently outputs action $a$. The action space consists of rotation about the yaw axis and displacement along the frontal axis in the robot frame, which allows all actions to be expressed in polar coordinates. As it is known that VLMs struggle to reason about continuous coordinates \cite{rahmanzadehgervi2024visionlanguagemodelsblind}, we instead transform the navigation problem into the selection of an action from a discrete set of options \cite{yang2023setofmark}. Our core idea is to choose these action options in a way that avoids obstacle collisions and promotes exploration.

Figure \ref{fig:overview} summarizes our approach. We start by determining the navigability of the local region by estimating the distance to obstacles using a depth image (Sec.~\ref{sec:mapupdate}). Similar to \cite{chang2023goat, shah2023lfg, exploreeqa2024, sathyamoorthy2024convoicontextawarenavigationusing, gadre2023cows, yu2023l3mvn, topiwala2018frontierbasedexplorationautonomous} we use the depth image and pose information to maintain a top-down voxel map of the scene, and notably mark voxels as \textit{explored} or \textit{unexplored}.
Such a map is used by an \textit{Action Proposer} (Sec. \ref{sec:actionproposer}) to determine a set of actions that avoid obstacles and promote exploration. We then project this set of possible actions to the first-person-view RGB image with the \textit{Projection} (Sec. \ref{sec:projection}) component. Finally, the VLM takes as input this image and a carefully crafted prompt, described in Sec.~\ref{sec:prompting}, to select an action, which the agent executes. To determine episode termination, we use a separate VLM call, detailed in Sec. \ref{sec:term}.

\subsection{Navigability}
\label{sec:mapupdate}

 \begin{wrapfigure}[15]{r}{0.5\textwidth}
    \vspace{-10pt}    
    \centering 

    \includegraphics[width=0.55\textwidth]{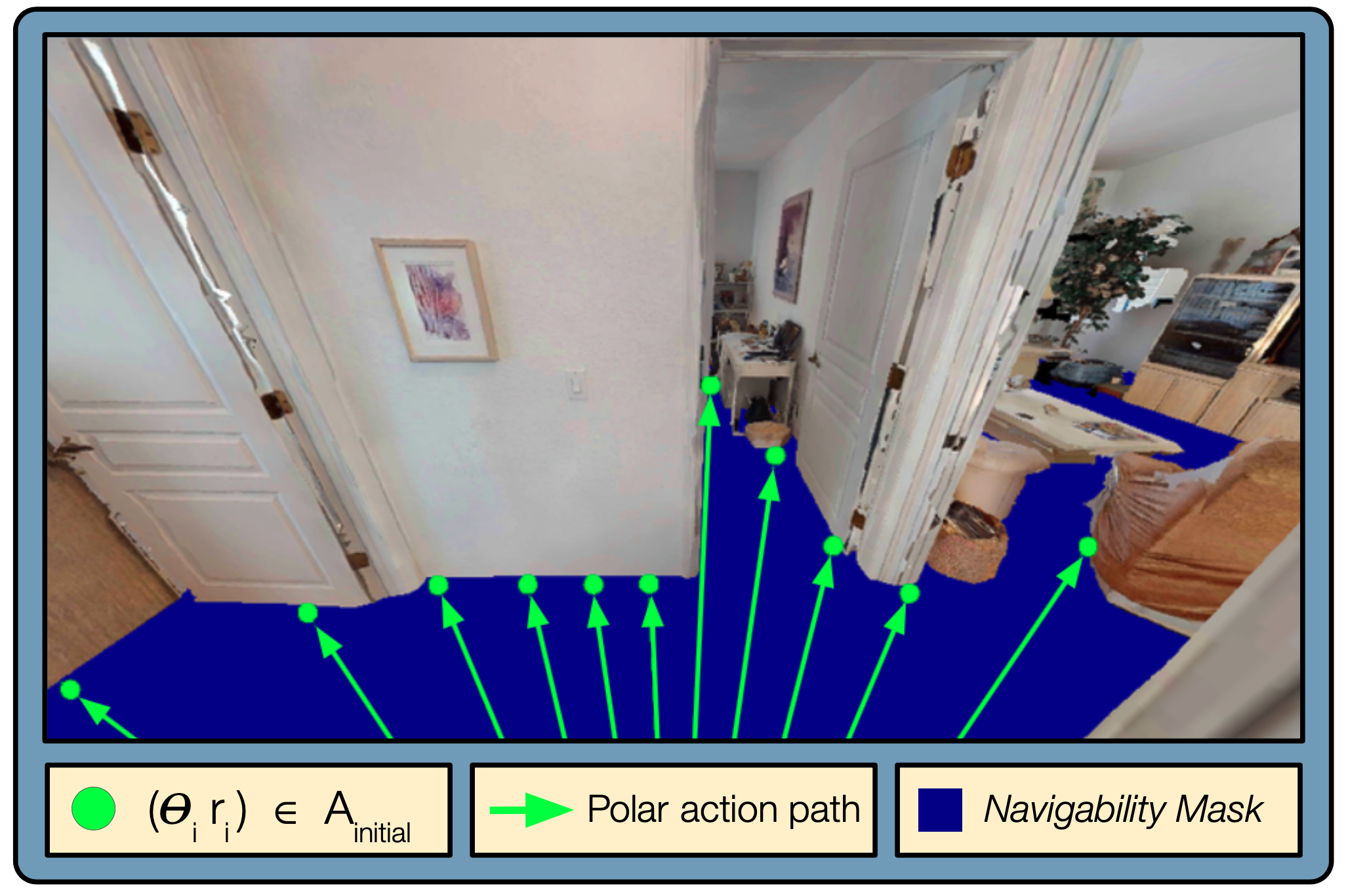}  

    \caption{\footnotesize An example step of the \textit{Navigability} subroutine. The navigability mask is shown in blue and polar actions making up $A_\text{initial}$ are in green}
    \label{fig:3}
\end{wrapfigure}

Using a depth image, we compute a \textit{navigability mask} that contains the set of pixels that can be reached by the robot without crashing into any obstacles. 

Next, for all directions $\theta \in fov$, we use the \textit{navigability mask} to calculate the farthest straight-line distance $r$ that the agent can travel without colliding. This creates a set of actions $A_\text{initial}$ that are collision-free. Figure \ref{fig:3} illustrates an example calculation of the mask and navigable actions.

At the same time, we use the depth image and the pose information to build a 2D voxel map of the environment. All observable areas within 2 meters of the agent are marked as \textit{explored}, and the ones beyond as \textit{unexplored}.

\subsection{Action Proposer}
\label{sec:actionproposer}
  
We design the \emph{Action Proposer} routine to refine  $A_\text{initial} \rightarrow A_\text{final}$, an action set that is interpretable for the VLM and promotes exploration.
Taking advantage of the information accumulated in our voxel map, we look at each action and define an exploration indicator variable $e_i$ as 
\[
e_i = 
\begin{cases} 
1 & \text{if region $(\theta_i, r_i)$ is unexplored} \\
0 & \text{if region $(\theta_i, r_i)$ is explored}
\end{cases}
\]
To build $A_\text{final}$, we need to prioritize unexplored actions, and also ensure there is enough visual spacing between actions for the VLM to discern. We start by adding unexplored actions to $A_{\text{final}}$ if an angular spacing of $\theta_\delta$ is maintained.

\[
A_{\text{final}} \gets A_{\text{final}} \cup \{(\theta_i, r_i) \mid e_i = 1 \text{ and } |\theta_i - \theta_j| \geq \theta_\delta, \forall (\theta_j, r_j) \in A_{\text{final}}\}
\]
To sufficiently cover all directions but still maintain an exploration bias, we supplement $A_{\text{final}}$ by adding explored actions subject to a \emph{larger} angular spacing of $\theta_\Delta > \theta_\delta$ :
\[
A_{\text{final}} \gets A_{\text{final}} \cup \{(\theta_i, r_i) \mid e_i = 0 \text{ and } |\theta_i - \theta_j| \geq \theta_\Delta, \forall (\theta_j, r_j) \in A_{\text{final}}\}
\]
Lastly, we want to ensure these actions don't move the agent too close to obstacles, so we clip \[r_i \gets \min(\frac{2}{3}\cdot r_i, \ r_{max}) \quad \forall (\theta_i, r_i) \in A_{\text{final}} \] 

Occasionally, the agent can get stuck in a corner where there are \textit{no} navigable actions ($A_\text{initial} = \emptyset$). To address this, we add a special action $(\pi, 0)$, which rotates the agent by 180\textdegree. This also allows efficient entry/exit of rooms where the agent quickly identifies that the goal is not in that room.

The proposed set $A_{\text{final}}$ now has three important properties: (i) actions correspond to navigable paths, (ii) there is sufficient visual spacing between actions, and (iii) there is an engineered bias towards exploration. We call this approach to exploration \textit{explore bias}.

\subsection{Projection}
\label{sec:projection}
Visually grounding these actions in a space the VLM can understand and reason about is the next step.
The \emph{Projection} component takes in $A_\text{final}$ from \ref{sec:actionproposer} and RGB image $I$, and outputs annotated image $\hat{I}$. Similarly to \cite{google2024pivot}, each action is assigned a number and overlayed onto the image. We assign the special rotation action with $0$ and annotate it onto the side of the image along with a label \textit{Turn Around}. We find that visually annotating it, instead of just describing it in the textual prompt, helps ground its probability of being chosen to that of the other actions.

\subsection{Prompting}
\label{sec:prompting}
To elicit a final action, we craft a detailed textual prompt $T$, which is fed into the VLM along with $\hat{I}$. This prompt primarily describes the details of the task, the navigation goal, and how to interpret the visual annotations. 
Additionally, we ask the model to describe the spatial layout of the image and to make a high-level plan \textit{before} choosing the action, which serves to 
improve reasoning quality as found by \cite{wei2023chainofthoughtpromptingelicitsreasoning, kojima2022large}. For image-based navigation goals, the goal image is simply passed into the VLM in addition to $T$ and $\hat{I}$. The full prompt can be found in Figure \ref{fig:1}. 

The action chosen by the VLM, $P_{\text{vlm}}(a^* | \hat{I}, T) \in A_\text{final}$ is then directly executed in the environment. Notably, this does not involve any low-level obstacle avoidance policy as in other works \cite{chang2023goat, shah2023lfg, gadre2023cows, yu2023l3mvn, kuang2024openfmnavopensetzeroshotobject}.

\subsection{Termination}
\label{sec:term}
\begin{figure}[h]
    \centering 
    \includegraphics[width=1\textwidth]{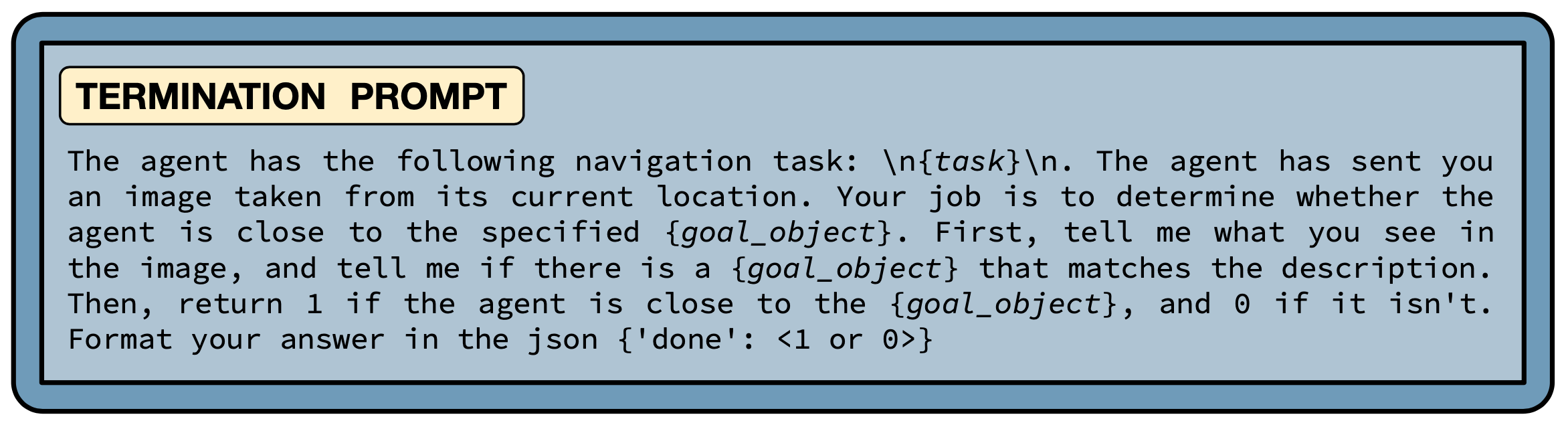}  
    \caption{\footnotesize The separate prompt for determining episode termination}
    \label{termination}
\end{figure}
To complete a navigation task, the agent must terminate the episode by calling special action \textit{stop} within a threshold distance of the goal object. Compared to other approaches that leverage a low-level navigation policy \cite{chang2023goat, shah2023lfg, gadre2023cows, yu2023l3mvn, kuang2024openfmnavopensetzeroshotobject}, our method does not explicitly choose a target coordinate location to navigate to, and therefore we face an additional challenge of determining when to stop. Our solution is to use a separate VLM prompt that explicitly asks whether or not to stop, which is shown in Figure \ref{termination}. We do this for two reasons:
\begin{enumerate}
    \item Annotations: The arrows and circles from Sec. \ref{sec:projection} introduce noise and clutter to the image, making it more difficult to understand.
    \item Separation of tasks. To avoid any task interference, the action call is only concerned with navigating and the stopping call is only concerned with stopping.
\end{enumerate}
To avoid terminating the episode too far away from the object, we terminate the episode when the VLM calls \textit{stop} two times in a row. After the VLM calls \textit{stop} the first time, we turn off the navigability and explore bias components to ensure the agent doesn't move away from the goal object.

\section{Experiments}
\label{sec:experiments}  

We evaluate our approach on two popular embodied navigation benchmarks, ObjectNav \cite{batra2020objectnavrevisitedevaluationembodied} and GoatBench \cite{khanna2024goatbench}, which use scenes from the Habitat-Matterport 3D dataset \cite{yadav2023habitatmatterport3dsemanticsdataset, savva2019habitatplatformembodiedai}.
Further, we analyze how the performance of an end-to-end VLM agent changes with variations in design parameters such as field-of-view, length of the contextual history used to prompt the model, and quality of depth perception. 

\textbf{Setup}:
Similar to \cite{habitatchallenge2022}, the agent adopts a cylindrical body of radius 0.17m and height 1.5m. We equip the agent with an egocentric RGB-D sensor with resolution (1080, 1920) and a horizontal field-of-view (FOV) of 131$^\circ$. The camera is tilted down with a pitch of 25$^\circ$ similar to \cite{exploreeqa2024}, which helps determine navigability. We use Gemini Flash as the VLM for all our experiments, given its low cost and high effectiveness.

\textbf{Metrics:} As in prior work \cite{khanna2024goatbench, habitatchallenge2022,anderson2018evaluationembodiednavigationagents}, we use the following metrics: (i) Success Rate (SR): fraction episodes that are successfully completed (ii) Success Rate Weighted by Inverse Path Length (SPL): a measure of path efficiency.

\begin{figure}[h]
    \centering 
    \includegraphics[width=1\textwidth]{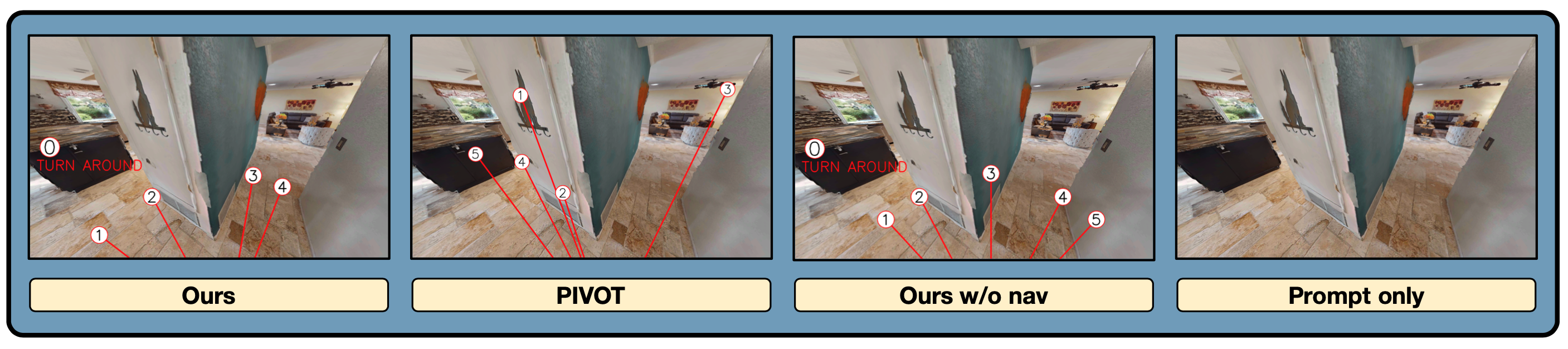}  
    \caption{\footnotesize\textbf{Baselines:} Comparing the four different methods on a sample image. \textit{Ours} contains arrows that point to navigable locations, \textit{PIVOT} has arrows sampled from a random 2-D Gaussian, \textit{Ours w/o nav} sees uniformly spaced arrows (note arrows 3 and 5 point into a wall), and \textit{Prompt Only} sees just the raw RGB image}
    \label{baselines}
\end{figure}

\textbf{Baselines:} We use PIVOT \cite{google2024pivot} as a baseline as it is most similar to ours. To investigate the impact of our action selection method, we ablate it by evaluating \textit{\baseline}: the same as ours but without the \textit{Navigability} and \textit{Action Proposer} components. The action choices for this baseline are a static set of evenly-spaced action choices, including the \textit{turn around} action. Notably, these actions do not consider navigability or exploration. To further evaluate the impact of visual annotation, we also evaluate a baseline \textit{Prompt Only}, which sees actions described in text (``turn around", ``turn right", ``move forward", ...) but not annotated visually. These different prompting baselines can be visualized in Fig \ref{baselines}.

We note that in our experiments and baselines, we turn the \textit{allow\_slide} parameter on, which allows the agent to slide against obstacles in the simulator. Our experiments show that removing this assumption leads to large drops in performance.

\subsection{ObjectNav}

The Habitat ObjectNav benchmark requires navigation to an object instance from one of six categories \textit{[Sofa, Toilet, TV, Plant, Chair, Bed]}. As in \cite{habitatchallenge2022}, to get the optimal path length, we take the minimum of the shortest paths to all instances of the object. These experiments are evaluated with a success threshold of 1.2 meters \cite{shah2023lfg}. 

\begin{table}[htbp]
\centering
\begin{minipage}{0.45\textwidth}
    \centering
    \begin{tabular}{|l|c|c|}
    \hline
    \textbf{Run}                     & \textbf{SR} & \textbf{SPL} \\
    \hline
    Ours & \textbf{50.4\%} & \textbf{0.210} \\
    \baseline & 33.2\% & 0.136 \\
    Prompt Only & 29.8\% & 0.107 \\
    PIVOT \cite{google2024pivot} & 24.6\% & 0.106 \\
    \textcolor{gray}{Ours w/o sliding} & \textcolor{gray}{12.9\%} & \textcolor{gray}{0.063} \\

    \hline
    \end{tabular}
\end{minipage}%
\hfill
\begin{minipage}{0.55\textwidth}
\vspace{7pt}  
{\caption{\footnotesize \textbf{ObjectNav Results.} We evaluate four different prompting strategies on the ObjectNav benchmark, and see our method achieves highest performance in both accuracy (SR) and efficiency (SPL). Ablating the \textit{allow\_slide} parameter shows our method is dependent on sliding past obstacles}
\label{tab:1}
}
\end{minipage}
\end{table}

Table \ref{tab:1} summarizes our results. Our method outperforms PIVOT by over $25\%$, and nearly doubles its navigation efficiency in terms of SPL. We see that our action selection method is highly effective as shows a $17\%$ improvement over \textit{Ours w/o nav}. Removing visual annotations leads to a slight decrease in success rate but a significant reduction in SPL, indicating that visual grounding is important for navigation efficiency. Interestingly, we find that PIVOT performs worse than both of our ablations. We attribute this to limited expressivity in its action space, which prevents it from executing large rotations or turning around fully. This often leads to the agent getting stuck in corners, hindering its ability to recover and navigate effectively.

We note that disabling sliding results in a large drop in performance, signaling that while effective in simulation, our method would likely lead to collisions with obstacles in the real world. While our \textit{Navigability} module can identify navigable locations, it does not consider the specific size and shape of the robot in this calculation, leading to occasional collisions where the agent gets stuck since we lack an explicit action to backtrack previous motions.

\subsection{Go To Anything Benchmark (GOAT)}
GOAT Bench \cite{khanna2024goatbench} is a recent benchmark that establishes a higher level of navigation difficulty. Each episode contains 5-10 sub-tasks across three different goal modalities: (i) Object names, such as \textit{refrigerator}, (ii) Object images, and (iii) Detailed text descriptions such as \textit{Grey couch located on the left side of the room, next to the picture and the pillow}. Table \ref{goat} shows our results, evaluated on the val unseen split.

\begin{table}[h]
\centering
\begin{adjustwidth}{1.2cm}{1.2cm} 
\begin{tabular}{|l|c|c|c|c|c|}
\hline
\textbf{Run}                     & \textbf{SR} & \textbf{SPL} & \textbf{Image SR} & \textbf{Object SR} & \textbf{Description SR} \\
\hline
Ours & \textbf{16.3}\% & \textbf{0.066}  & \textbf{14.3}\% & \textbf{20.5}\% & \textbf{13.4}\% \\
\baseline & 11.8\% & 0.054 & 7.8\% & 16.5\% & 10.2\% \\
Prompt Only & 11.3\% & 0.037 & 7.7\% & 15.6\% & 10.1\% \\
PIVOT \cite{google2024pivot} & 8.3\% & 0.038 & 7.0\% & 11.3\% & 5.9\% \\
\hline
\end{tabular}
\vspace{5pt}
\caption{\footnotesize{\textbf{GOAT Results.} Comparison of prompting strategies on GOAT Bench, a more challenging navigation task. Across three different goal modalities, our method strongly outperforms baseline methods}}
\vspace{-10pt}  
\label{goat}
\end{adjustwidth}
\end{table}

Across all goal modalities, our model achieves significant improvements over baselines. These improvements are especially evident in image goals, where our model achieves nearly twice the success rate of all baseline methods. This highlights the robustness and general nature of our system. As with the ObjectNav results, \textit{Ours w/o nav} and \textit{Prompt only} perform comparable, and both outperform PIVOT. For all prompting methods, the image and description modalities prove more challenging than the object modality, similarly to what was found by \cite{khanna2024goatbench}.

\textbf{Comparison to state-of-the-art:} We turn the \textit{allow\_slide} parameter off and compare to two state-of-the-art specialized approaches: (i) SenseAct-NN~\cite{khanna2024goatbench} is a policy trained with reinforcement learning, using learned submodules for different skills; and (ii) Modular GOAT~\cite{chang2023goat} is a compound system that builds a semantic memory map of the environment and uses a low-level policy to navigate to objects within this map. Unlike SenseAct-NN, our work is zero-shot, and unlike Modular GOAT, we do not rely on a low-level policy or a separate object-detection module.
\begin{table}[h]
\centering
\begin{minipage}{0.55\textwidth}
    \centering
    \begin{tabular}{|l|c|c|}
    \hline
    \textbf{Run}  & \textbf{SR} & \textbf{SPL} \\
    \hline
    SenseAct-NN Skill Chain & \textbf{29.5\%} & 0.113 \\
    Modular GOAT & 24.9\% & \textbf{0.172} \\
    \textcolor{gray}{Ours w/ sliding} & \textcolor{gray}{16.3\%} & \textcolor{gray}{0.066} \\
    Ours & 6.9\% & 0.049 \\
    \hline
    \end{tabular}
\end{minipage}%
\begin{minipage}{0.45\textwidth}
    \caption{\footnotesize{Directly comparing to other works, we see that specialized systems still produce superior performance. We also note these other works use a narrower FOV, lower image resolution, and a different action space, which could explain some of the differences}}
    \label{GOAT_SOTA}
\end{minipage}
\end{table}

We compare the results of our approach to these baselines in Table~\ref{GOAT_SOTA}. Interestingly, these methods have different strengths: a reinforcement learning approach leads to the highest success rate. Conversely, the modular navigation system achieves the highest navigation efficiency.

Our method shows lower performance compared to these specialized baselines across both metrics, even when permitted to slide over obstacles. Notably, we observe that in $13.9\%$ of the runs, the VLM prematurely calls \textit{stop} when it is between 1 to 1.5 meters from the target object. These instances are classified as failures, as the benchmark defines a run as successful only if the agent is within 1 meter of the object. This finding suggests that our VLM lacks the fine-grained spatial awareness necessary to accurately assess distances to objects. However, it also indicates that in over 30\% of the runs, our VLM agent is able to approach the goal object closely, highlighting its capability to reach near-target positions.

As shown in previous experiments, when not allowed to slide over objects, our approach's performance drastically decreases, as it gets frequently blocked between obstacles and does not have a way to backtrack its actions.

\subsection{Exploring the design space of VLM agents for navigation}
In this section, we look at major design choices that impact the navigation ability of VLM-based agents in our setup, all evaluated on the ObjectNav dataset.

\subsubsection{How important is camera FOV for navigation?}

\begin{figure}[h]
    \centering 
    \begin{adjustwidth}{0.6cm}{0.7cm} 

    \includegraphics[width=0.9\textwidth]{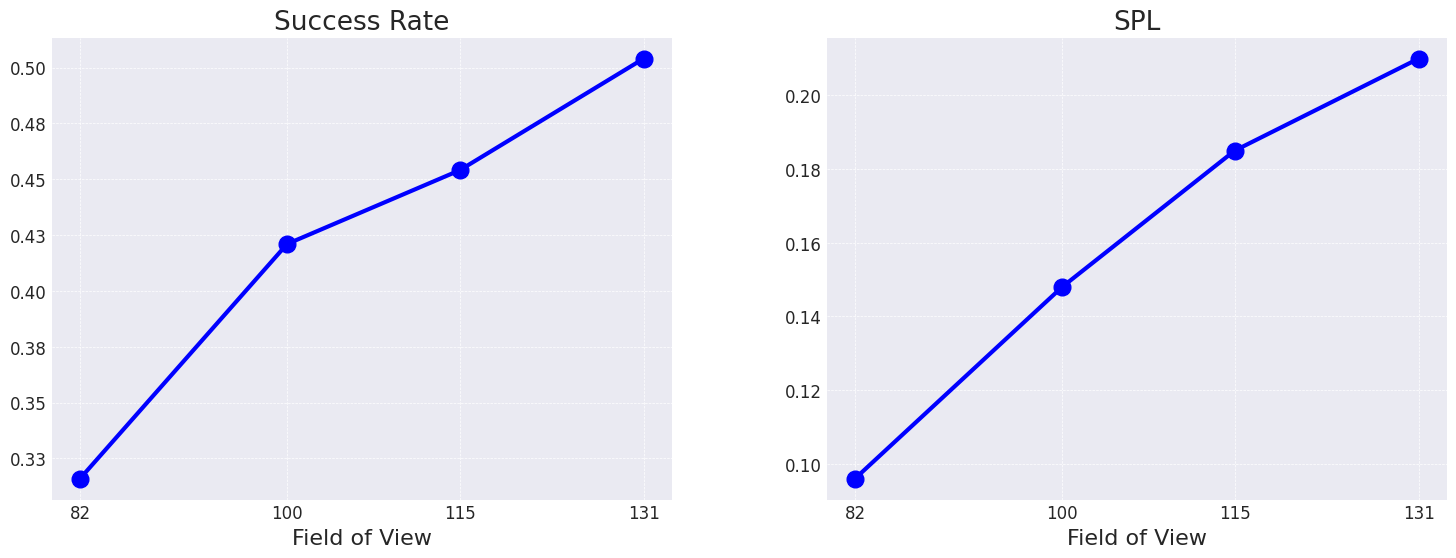}  
    \caption{\footnotesize\textbf{Impact of sensor FOVs.} We evaluate the performance of four different sensor FOVs, and find that a wider FOV invariably leads to higher performance}
    \label{fig:4}
    \end{adjustwidth}
\end{figure}

An agent's navigation abilities largely depend on how fine-grained its vision is. In this section, we study whether our VLM agent can benefit from high-resolution images. Specifically, we run our method using four different FOVs: 82$^\circ$ \cite{habitatchallenge2022}, 100$^\circ$, 115$^\circ$ and 131$^\circ$ (iPhone 0.5 camera). 
The results of this experiment, shown in Fig.~\ref{fig:4}, indicate positive scaling behaviors on both navigation accuracy and efficiency.

\subsubsection{Do longer observation-action histories help?}
In this section, we study whether a VLM navigation agent can effectively use a history of observations. We create a prompt containing the observation history in a naive way, i.e., we concatenate observations and actions from the $K$ most recent environment steps and feed this into the VLM as context. For all these experiments, we remove our exploration bias (see Sec.~\ref{sec:actionproposer}) to specifically isolate the contribution of a longer history.

\begin{table}[htbp]
\centering
\begin{minipage}{0.44\textwidth}
    \centering
    \begin{tabular}{|c|c|c|}
    \hline
    \textbf{History Length}                     & \textbf{SR} & \textbf{SPL} \\
    \hline
    No history & 46.8\% & 0.193 \\
    5  & 42.7\% &  0.180      \\
    10  & 45.4\% &  0.196    \\
    15  & 40.4\% &  0.170       \\
    \hline
    \end{tabular}
\end{minipage}%
\hfill
\begin{minipage}{0.56\textwidth}
    \caption{\footnotesize \textbf{Impact of adding context history.} We compare our method to alternatives of keeping the past 0, 5, 10, and 15 observations and actions. We see that adding context history does not improve the performance of our method}
    \label{memory}
\end{minipage}
\end{table}

The results of these experiments are shown in Table \ref{memory}. We find that when naively concatenating past observations and actions, our prompt strategy is unable to use a longer context. Indeed, the performance remains the same or decreases when increasing the history length.

\subsubsection{How important is perfect depth perception?}

Within the simulator, the depth sensor provides accurate pixel-wise depth information, which is important for determining the navigability mask. To investigate the importance of quasi-perfect depth perception, we evaluate two alternate approaches that only use RGB: (i) \textbf{Segformer}, which uses \cite{xie2021segformersimpleefficientdesign} to semantically segment pixels belonging to the \textit{floor} region. We use this region as the \textit{navigability mask} and bypass the need for any depth information. We estimate the distances to obstacles by multiplying the number of pixels with a constant factor. (ii) \textbf{ZoeDepth}, which uses \cite{bhat2023zoedepthzeroshottransfercombining} to estimate metric depth values. We use such predicted values instead of the ground-truth distances from the simulator and compute navigability in the original way. 

\begin{table}[h]
\begin{minipage}{0.5\textwidth}
    \centering
\centering
\begin{tabular}{|l|c|c|}
\hline
\label{Table 4}
\textbf{Run}                     & \textbf{SR} & \textbf{SPL} \\
\hline
Depth sensor  & 50.4\% & 0.210  \\
Segformer \cite{xie2021segformersimpleefficientdesign}  & 47.2\% & 0.183 \\
ZoeDepth \cite{bhat2023zoedepthzeroshottransfercombining}  & 39.1\% & 0.161 \\
\hline
\end{tabular}
\end{minipage}
\hfill
\begin{minipage}{0.5\textwidth}
\centering
\caption{\footnotesize \textbf{Depth Ablation.} We evaluate two alternate approaches that only require RGB. We find that semantic segmentation performs close to using ground truth depth, whereas estimating depth values leads to a significant performance drop \label{tab:rgb_only} }
   
\end{minipage}
\end{table}

The results of this study are presented in Table~\ref{tab:rgb_only}. We find that depth estimation from \cite{bhat2023zoedepthzeroshottransfercombining} is not accurate enough to identify navigable areas. Indeed, depth noise leads to a $10\%$ drop in SR. However, using a segmentation mask instead of relying on depth information surprisingly proves to be quite effective, with only a decrease of $3\%$ with respect to using perfect depth perception. Overall, our experiments show that a VLM-based navigation agent can perform well with only RGB information.

\section{Conclusion}
\label{sec:conclusion}
In this work, we present VLMnav, a novel visual prompt-engineering approach that enables an off-the-shelf VLM to act as an end-to-end navigation policy. The main idea behind this approach is to carefully select action proposals and project them on an image, effectively transforming the problem of navigation into one of question-answering. Through evaluations on the ObjectNav and GOAT benchmarks, we see significant performance gains over the iterative baseline PIVOT, which was the previous state-of-the-art in prompt engineering for visual navigation. Our design study further highlights the importance of a wide field of view and the possibility of deploying our approach with minimal sensing, i.e., only an RGB image.

Our method has a few limitations. The drastic decrease in performance from disabling the \textit{allow\_slide} parameter indicates that there are several collisions with obstacles, which could be problematic in a real-world deployment. In addition, we find that specialized systems such as \cite{khanna2024goatbench} outperform our work. However, as the capabilities of VLMs continue to improve, we hypothesize that our approach could help future VLMs reach or surpass the performance of specialized systems for embodied tasks.

\bibliography{example}  

\end{document}